# Automatic phantom test pattern classification through transfer learning with deep neural networks

Rafael B. Fricks[a,b], Justin Solomon[b], Ehsan Samei[b]

[a]Department of Veterans Affairs, Durham VA Medical Center, 508 Fulton St, Durham, NC, USA 27705; [b]Duke University, Carl E. Ravin Advanced Imaging Laboratories, 2424 Erwin Rd, Durham, NC, USA 27705

## ABSTRACT

Imaging phantoms are test patterns used to measure image quality in computer tomography (CT) systems. A new phantom platform (Mercury Phantom, Gammex) provides test patterns for estimating the task transfer function (TTF) or noise power spectrum (NPF) and simulates different patient sizes. Determining which image slices are suitable for analysis currently requires manual annotation of these patterns by an expert, as subtle defects may make an image unsuitable for measurement. We propose a method of automatically classifying these test patterns in a series of phantom images using deep learning techniques. By adapting a convolutional neural network based on the VGG19 architecture with weights trained on ImageNet, we use transfer learning to produce a classifier for this domain. The classifier is trained and evaluated with over 3,500 phantom images acquired at a university medical center. Input channels for color images are successfully adapted to convey contextual information for phantom images. A series of ablation studies are employed to verify design aspects of the classifier and evaluate its performance under varying training conditions. Our solution makes extensive use of image augmentation to produce a classifier that accurately classifies typical phantom images with 98% accuracy, while maintaining as much as 86% accuracy when the phantom is improperly imaged.

## 1. INTRODUCTION

The Mercury phantom (Gammex) is a test object for assuring image quality by measuring properties of a commercial computed tomography (CT) system such as its task transfer function (TTF) or noise power spectrum (NPS) (Figure 1A)[1]. It provides embedded test patterns for quantifying these imaging system properties in scanners (Figure 1B). Currently, using the phantom requires an operator to manually annotate the locations of test patterns in a series of axially acquired images. While some patterns are easily distinguished, manual annotation is still required as subtle distinctions may make an image unsuitable (Figure 2). Manual annotations are time-consuming, suffer from inter-reader variability, and may lead to discarding useable phantom images. This study aims to automate the image annotation process using deep learning methods for image classification.

A variety of computer vision tasks have been successfully automated via deep learning approaches[2–12]. The effectiveness of a deep learning approach to a novel problem is typically limited by the data and computational power available. Transfer learning techniques greatly accelerate image classifier training by reusing pretrained neural networks[11,13]. Pretrained classifiers such as VGG19[12] benefit from training on over 14 million natural images in ImageNet[14]. Phantom imaging introduces unique considerations for transfer learning. The natural images in ImageNet are typically uncorrelated images, photographed with three channels of color information. Unlike natural images, the phantom images acquired through CT are monochrome. The serial acquisitions produce highly correlated neighboring images, and indeed experts rely on local context in labeling the slices. For instance, tapered sections are composed of the same homogenous material as NPS measurement patterns, however unlike tapered sections the NPS section diameter remains constant for the length of an NPS test pattern segment (Figure 1). Similarly, air gaps and other artifacts that make test patterns unsuitable tend to occur at material interfaces. Phantom images do not require color channels, but local context is informative.

Another distinction from natural images is that by design the phantom is an invariant imaging target composed of distinct geometric shapes. Minor differences in imaging conditions introduce variations—such as phantom position and alignment or reconstructed field-of-view—rather than the imaged object. Differences in placing the phantom within the CT will axially shift the location and severity of air gap helical artifacts at different material interfaces. Misaligning the phantom with the scanner may rotate or warp slices in the imaging plane. Finally, changes in the reconstructed field of view affects image scale. These variations may hypothetically be emulated by image augmentation entirely, without the need for physically acquiring more representative data.

This work aims to produce a classifier that is insensitive to misalignments but consistent in reporting air gaps or partial volumes. Our proposed solution appends one image in either direction as a three-channel image, allowing for effective re-use of well-established ImageNet classification networks trained on three-channel color images. Extensive augmentation is employed to train a network to generalize for a variety of imaging conditions. This study adapts training techniques for transfer learning with imaging phantoms, using the Mercury Phantom V3.0[1] as an example.

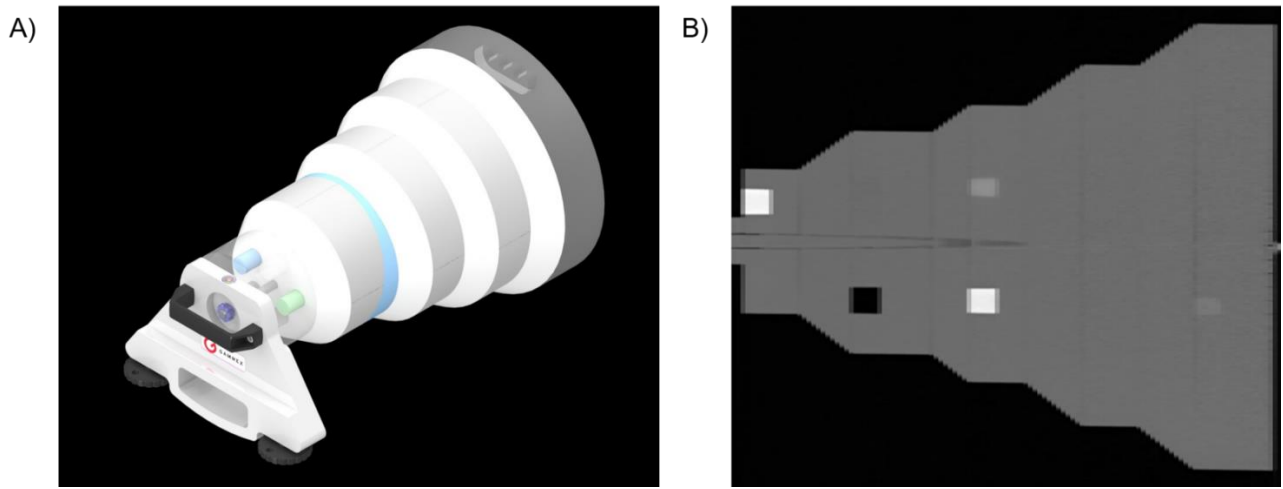

Figure 1. (A) Design rendering of a Mercury Phantom with positioning equipment. (B) Sagittal reconstruction of a series of Mercury Phantom images, aspect ratio adjusted for visualization. Around 114 images are acquired at 5mm slice thickness using helical CT, with images reconstructed to show approximately 400 mm field-of-view at 512x512 pixel resolution.

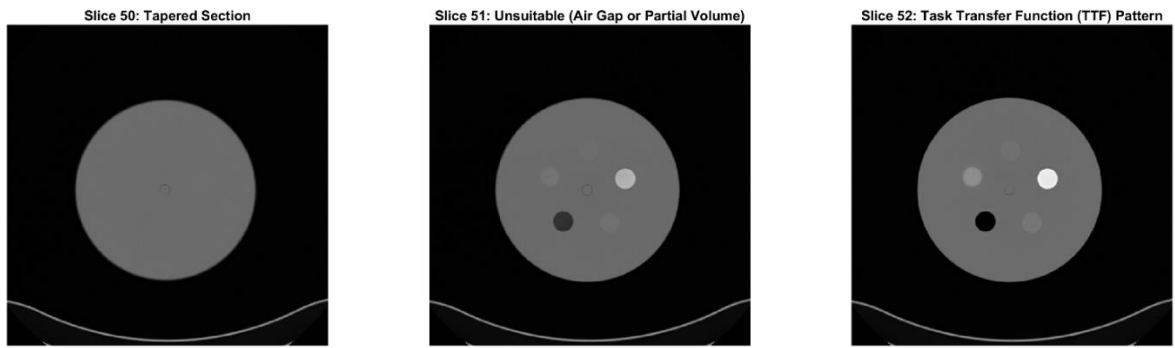

Figure 2. Axial images from a series of phantom images. Slice 50 has homogenous interior, consistent with a noise power spectrum (NPS) estimation pattern but is part of a tapered section of the phantom which serves a different purpose. Slice 51 captures a transition from the tapered section to the task transfer function (TTF) pattern, causing faint blending in the test pattern, and thus making it unsuitable for TTF measurements. Slice 52 is a suitable TTF pattern image. Local axial context is critical for human observers in making these assessments.

## 2. METHODS

### 2.1 Phantom Data

Data for this study were drawn from testing patterns at a university medical center. Fifteen image series of the Mercury Phantom V3.0[1] were drawn from routine clinical physics testing data and represented images from a variety of scanner models and at variable radiation dose levels. Annotating these series yielded 3099 images of phantom slices. We divided the images at the series level into nine series for training (2409 images), three for a validation set (347 images), and three for a test set (343 images). These sets represent a range of typical phantom imaging conditions. These images were acquired at 400mm field-of-view, with slice thicknesses of 5mm. Each image in a series is labeled according to one of five classes; 1) Outside Phantom, 2) Noise Power Spectrum (NPS) pattern, 3) Task Transfer Function (TTF) pattern, 4) Tapered Section, 5) Unsuitable (Air gap or partial volume present). Figure A1 depicts labeled images in order from the training set, with no shuffling or augmentation (See Appendix).

An additional series was acquired at atypical conditions, yielding 458 annotated images of a mercury phantom at oblique angles. We refer to this series as the extreme test. This series is beyond the recommended use of the phantom and serves as an evaluation set for algorithm performance under misalignment (See Appendix, Figure A2).

### 2.2 Preprocessing, Augmentation, and Transfer Learning

In preprocessing, we windowed the CT images into the empirically determined range of [-1024, 1187] Hounsfield Units. Images were down-sampled from 16-bit unsigned integer pixel intensities at 512x512 reconstructed resolution, to 8-bit unsigned integers at 256x256 resolution. Each image was converted to a three-channel image by appending the previous and next image in the series axially. For images at the edge of the acquisition, the central image was repeated as necessary, i.e. the first image used a copy of the first image as the 'previous' image. The three-channel appended format added axial context, and each resulting input was labeled according to the central image.

To emulate misalignments during training, we augmented the images extensively at runtime. Augmentations included vertical flipping, horizontal flipping, applying rotations up to 90 degrees, magnification from 90%-110% image size, horizontal and vertical translations up 20% of the image size. Additional augmentations such as a random brightness of up to 0.5% of pixel intensity, and motion blurring using kernels between 3x3 and 7x7 were also applied randomly. Each of these augmentation effects were enabled at runtime on each input with 50% probability and applied by a custom image generator using the Albumentations package[15]. Augmentations are applied to the training set only.

We employed the VGG19 architecture[12] with ImageNet weights[14] for convolutional filters as a starting point. The output classification layers were replaced by a two-layer perceptron of two dense layers with $n_p$ channels each, followed by batch normalization[16] prior to ReLU activation. Dense layer weight parameters were randomly initialized using He initialization. The number of channels in each dense layer $n_p$ was varied across the range [256, 512, 1024, 2048, 4096] to find an optimal choice. The perceptron output feeds into a softmax activation for the final prediction, as pictured in Figure 3.

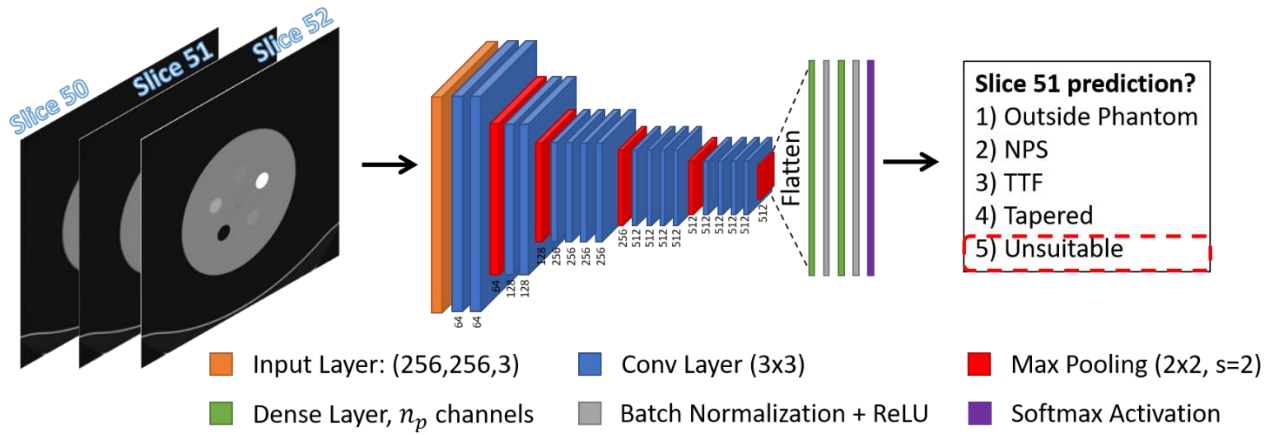

Figure 3. Three sequential images are down-sampled and concatenated to produce the input. A series of convolution and max pooling layers following the VGG19 architecture are applied to this input, with initial weights trained on ImageNet. Convolutional layers are padded to return a same-size output. The final max pooling layer output is flattened, then passed through added dense layers with batch normalization and ReLU activation. Lastly, applying a softmax activation yields a classification prediction for the central image in the input.

The model was trained for 1000 epochs at a batch size of 32. We used an Adam optimizer[17] (learning rate 0.001, $\beta_1 = 0.9$, $\beta_2 = 0.999$), epsilon of $10^{-8}$, with decay of 0.001. Categorical cross-entropy served as the loss function. The model was generated and trained using Keras model specification in Tensorflow[18] 2.0.0. We seeded random number generators identically at the start of training for each classifier. Training and evaluation used a single NVIDIA Titan RTX GPU, and took approximately 17 hours per model variant. Example code is available at (bit.ly/PC-SPIE2020).

### 2.3 Classifier Evaluation

At the conclusion of training, the model was used to classify the validation set, test set, and extreme test set. In each case we examined the misclassification or confusion matrix, as well as evaluated the precision, recall, and F1 score for each classification category. An overall class-weighted accuracy value is also calculated and reported. These evaluation results were used to select an optimal $n_p$ for this application, balancing classification performance and model size.

### 2.4 Ablation Experiments

Several design choices reflect our understanding of the unique challenges in this phantom imaging task. We performed a set of ablation experiments which modify the model to verify the effectiveness of several training decisions. Except where noted, in each scenario the classification model and training conditions were identical to the baseline result from evaluation, with $n_p$ fixed to the optimum determined in the evaluation step. The performance metrics (precision, recall, and F1 score) were evaluated on validation, test, and extreme test data sets as before.

#### 2.4.1 Transfer learning and random initialization

Given the dissimilarities between the photographs in ImageNet and x-ray computed tomography images of the Mercury Phantom, transfer learning with a natural image classifier may be less impactful than expected. We evaluated the utility of transfer learning from ImageNet weights by comparing training performance to an identical model where convolutional layer weights are randomly initialized according to a normal distribution with zero mean and unit variance.

#### 2.4.2 Axial information through channels

We assert that some regional information in the z direction, referred to as axial information, contributes to classification accuracy. This axial information is provided to the classifier at runtime by appending a previous and next image to the classification target, generating the three-channel input. To verify this assertion, we trained four variants of the classifier, progressively freezing convolutional layer parameters and preventing training up to the second max pooling layer. These variants are denoted [f0, f1, … f4], where f0 has no frozen layers (all parameters are trainable as in the baseline model), f1 has the first convolutional layer frozen, and f4 has the first four convolutional layers frozen and untrainable.

Additionally, we trained a variant of the f0 or baseline model where the same image is replicated in triplicate. Extending the example in Figure 3, the triplicate classifier receives three copies of Slice 51 data concatenated and makes a prediction for Slice 51 based on this input. Triplicating the input maintains the same overall number of trainable parameters as in the baseline case.

#### 2.4.3 Augmentation to emulate misalignment

Presuming the variability between imaging sets arises from positioning rather than changes in the mercury phantom, we apply augmentation to emulate suboptimal positioning in a well-positioned training set. We compared the classification accuracy on the extreme test set of the baseline model to a model trained without augmentation.

## 3 RESULTS

We evaluate the model performance for training, validation, and test sets as the number of connected layer parameters is varied. We also evaluate the model effectiveness at classifying images from the extreme test acquisition series. We then report results from ablation experiments to probe the ability of the classifier to 1) benefit from transfer learning, 2) incorporate axial data, and 3) learn to generalize from augmentation.

### 3.1 Classifier Evaluation

Examining the classification accuracy on the validation set at each epoch (Figure 4), all five model variants learn to adequately classify the phantom slices in the validation set over the course of training. The overlaid plots show all models converge at approximately 98% accuracy on this set. Similar results are seen when evaluating the classification results post-training for the validation set (see Appendix, Table A1). All model variants classify the validation set with 98% accuracy, except for the 4096 variant which only achieves 97% accuracy. This high parameter count also seems to cause mild overfitting, indicated by the larger fluctuations in classification accuracy from one epoch to another (Figure 4). The same fluctuation is evident in the loss plot, although smaller. There is overall no distinguishing difference in performance on this set, which indicates that $n_p = 256$ is the optimal choice as it significantly minimizes the model size (Table 1).

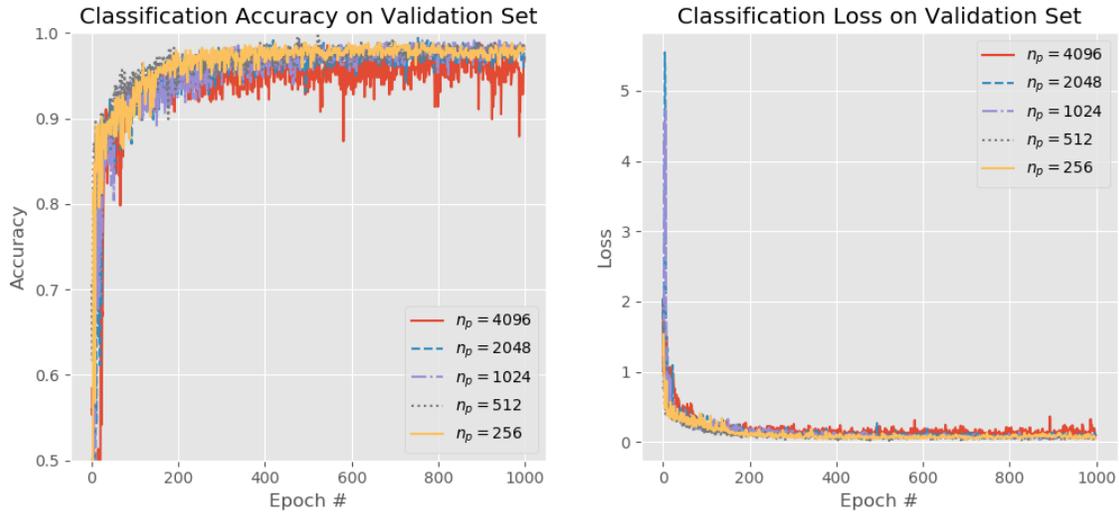

Figure 4. Model accuracy and loss on validation set, evaluated at the end of each training epoch. Lines are plotted in reverse order of model size, with $n_p = 4096$ first, overlaid by progressively smaller parameter values. All models appear to achieve the same validation set accuracy after approximately 400 epochs, regardless of the number of parameters in the dense layers. Models with significantly more parameters appear to experience more overfitting, as seen by the large variances in accuracy in the 4096 model through the epochs.

Classifying the test set with the trained model yields similar results as the validation set (Appendix Table A2). This is to be expected as the test and validation sets represent similar imaging conditions of a relatively invariant imaging target. All models achieve 97% accuracy, except for the 256 variant which now notably achieves 99% accuracy. There is no overall distinguishing performance.

Finally, applying the model to the extreme test set shows a notable decrease in classification accuracy (Appendix Table A3). Variants with $n_p = 1024, 2048$ do not achieve 80% accuracy in this evaluation. The 256 variant reaches similar accuracy as the 512 and 4096 variants (85% to 86% in the other two). Examining the confusion matrix closely, however, we see a distinct pattern separating the 256 from the higher-parameter variants. The 256 erroneously places several air gap or partial volume images (AGP) into the usable image categories. 512 and 4096 tend to balance errors with respect to the AGP category better, reducing the overall number of unsuitable images recommended. The 512 variant provides overall consistent performance with per-category precision and recall above 0.80 in most cases. The 512 model maximizes the F1 score with the exception of AGP, where the 4096 variant is slightly more effective.

| nP | Total Number of Parameters | Parameters in Dense Layers | % Total |
|---|---|---|---|
| 256 | 28,481,861 | 8,454,144 | 0.2968 |
| 512 | 37,070,405 | 17,039,360 | 0.4596 |
| 1024 | 54,640,709 | 34,603,008 | 0.6333 |
| 2048 | 91,354,181 | 71,303,168 | 0.7805 |
| 4096 | 171,072,581 | 150,994,944 | 0.8826 |

Table 1. Total number of parameters as the number of channels in the dense layers ($n_p$/nP) is changed. The first dense layer especially makes a multifold connection with every convolutional output of the final pooling layer. As the number of channels is increased, the share of total parameters devoted to the classification layers becomes a majority.

In current practice it is not necessary to use all available phantom images to evaluate the CT system. Multiple slices provide additional or perhaps redundant measurements. Therefore, suggesting an unsuitable image for use is a far less desirable error in this application than erroneously discarding a suitable image. We elect to use dense layers with 512 channels in all subsequent tests. It minimizes parameters to the second lowest level studied, while achieving a favorable error profile comparable to the model with the most parameters. The effect of this parameter on overall model size can be significant (Table 1). The extreme test performance proved far more informative than test or validation sets, where all models classify standard images reasonably well.

### 3.2 Ablation experiments

We proceed with ablation experiments based off the model with $n_p = 512$. All variants use the architecture in Figure 3, and identical procedures for training, except where noted. We highlight the most informative results of these experiments. All experiments are compared to a baseline control classifier, which is the unmodified 512-variant in the previous section.

#### 3.2.1 Transfer learning and random initialization

Randomizing the VGG19 parameters, rather than using weights trained on ImageNet, results in a delay in achieving the previously reported performance maxima. The randomly initialized model (RI) converges approximately with the pretrained model (Control) in validation test accuracy after 600 epochs (Figure 5). A slight decrease of 1% accuracy is noted in validation and test sets which are classified correctly in 97% and 96% of cases, respectively. The extreme test is more informative, showing more substantial accuracy losses (Table 2). ImageNet weights improve accuracy in the misaligned phantoms tested. Under normal imaging conditions the accuracy difference is indiscernible, although fewer epochs are required to attain similar accuracy.

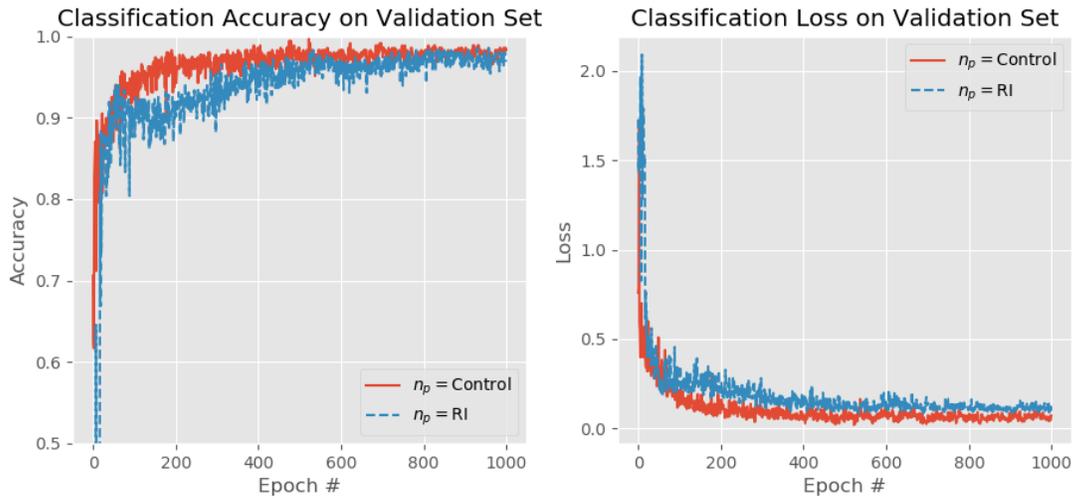

Figure 5. Model accuracy and loss on validation set, evaluated at the end of each training epoch. Here the control model represents the 512 variant model evaluated previously. In the RI model all parameters and initialization are identical to the control, except for the VGG19 convolutional layer parameters. Rather than loading pre-trained ImageNet weights, weights are randomly initialized according to a normal distribution with zero mean and unit variance. RI model accuracy nearly converges with the control by 600 epochs, with a slight gap remaining visible after 1000 training epochs in the loss plot.

| **Extreme Test Set (RI)** | **Confusion Matrix** | | | | | | **Metrics** | | |
|---|---|---|---|---|---|---|---|---|---|
| | Prediction | | | | | N Samples | Precision | Recall | F1 |
| Air Gap or Partial Volume (AGP) | 126 | 46 | 0 | 24 | 9 | 205 | 0.77 | 0.61 | 0.68 |
| Noise Power Spectrum (NPS) | 5 | 74 | 0 | 0 | 0 | 79 | 0.57 | 0.94 | 0.71 |
| Outside Phantom (OoP) | 3 | 0 | 1 | 0 | 0 | 4 | 1.00 | 0.25 | 0.40 |
| Task Transfer Function (TTF) | 3 | 0 | 0 | 55 | 0 | 58 | 0.70 | 0.95 | 0.80 |
| Tapered Section (TaS) | 26 | 10 | 0 | 0 | 76 | 112 | 0.89 | 0.68 | 0.77 |
| | AGP | NPS | OoP | TTF | TaS | TOT = 458 | Accuracy = 0.72 | | |

Table 2. The randomly initialized model underperforms on the extreme test set, notably failing to mark images with an air gap or partial volume as unsuitable images (AGP). Overall performance on validation and test sets, which are similar to training sets, is unremarkable from the evaluation variant, and not pictured here. The model is accurate on phantoms with more optimal placement, and errors follow the AGP misclassification pattern seen previously.

### 3.2.2 Axial information through channels

We employed two methods to examine the model's ability to incorporate regional information in the axial direction. In the first method several variants of the model were trained with the first n convolutional layers frozen. In model fn, the nth convolutional layer and all subsequent layers' weight values were not updated in training. The baseline model f0 denotes a model with no frozen layers and serves as a control. Model f1 freezes the first layer, f2 the first and second, and so on until f4, which freezes the first four convolutional layer parameters from training. All convolutional layers are initialized with weights trained on ImageNet.

Freezing these early layers has little discernible impact on the accuracy of this classifier. A mild effect is noted by examining the distribution of training accuracy over the second half of training (Figure 6). Freezing initial layers progressively results in apparent underfitting of the training cases which is only detectible as a drift of the mean accuracy when evaluated against the training set. Accuracy drifts downwards by approximately 0.5% in classifying the training set as more layers are withheld from training. We detect no significant impact on accuracy in classifying any independent evaluation sets compared to baseline.

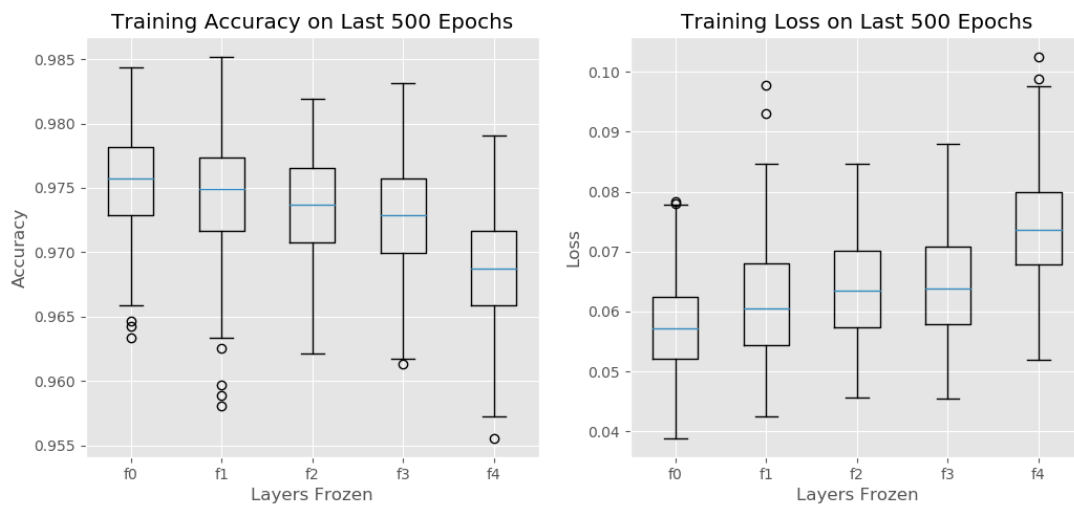

Figure 6. Box plots of accuracy and loss on the second half (last 500 epochs) of training, evaluated in classifying the training set at the end of each epoch. While all variations vary across approximately the same range of 98.5% accuracy to 96% accuracy, the mean accuracy steadily declines as layers are frozen. Evaluation sets show no change in accuracy.

As a second method of investigating how the model incorporates axial information, a triplicate model was trained which received replicas of the target image as input. The triplicate model has the same parameter count, but less potential input information. In training we clearly see a persistent reduction of approximately 2% validation set accuracy (Figure 7).

The accuracy loss in the triplicate model is consistent across all normal imaging conditions. Validation and test sets lose 2% and 3% overall accuracy respectively (Table 3). Interestingly the types of errors made are only related to image unsuitability (AGP). Losing axial information exacerbates the primary mode of classification failure the baseline model experienced. Expert labelers look for air gaps or partial volumes near material interfaces. Axial information appears to help in classifying these subtle defects.

More pronounced effects are seen again in the extreme test set. The model struggles to label unsuitable AGP images when classifying with triplicates. Also in this set a new mode of failure becomes apparent; the missing axial context causes tapered sections to be incorrectly classified as NPS test patterns in the extreme test. NPS segments in the mercury phantom are composed of the same material as TaS but maintain a constant diameter for the length of each NPS segment. When presented with axial images there is a detectible progression of phantom diameters that indicates a tapered section. Similarly, there is a consistent order to the test patterns; TTF patterns precede NPS segments, which precede TaS segments, which are followed by a new TTF segment. These associations are lost without the preceding and succeeding images. Losing key axial information appears to cause misclassifications of TaS slices as NPS patterns.

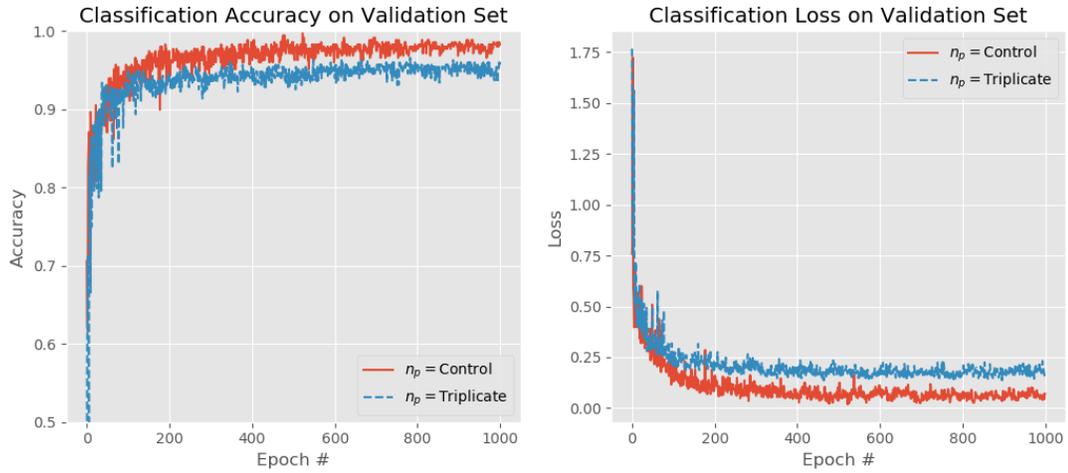

Figure 7. Comparison of the original model (control) to a model that receives the same image in triplicate. Plots show model accuracy and loss on validation set, evaluated at the end of each training epoch. While both models converge to a steady accuracy range, there is a discernible offset of approximately 2% accuracy loss when axial information is not available.

**Validation Set (Triplicate)**

| | \multicolumn{5}{c}{Confusion Matrix} | N Samples | Precision | Recall | F1 |
|---|---|---|---|---|---|---|---|---|---|
| | \multicolumn{5}{c}{Prediction} | | | | |
| Air Gap or Partial Volume (AGP) | 69 | 0 | 0 | 2 | 2 | 73 | 0.86 | 0.95 | 0.90 |
| Noise Power Spectrum (NPS) | 2 | 84 | 0 | 0 | 0 | 86 | 1.00 | 0.98 | 0.99 |
| Outside Phantom (OoP) | 0 | 0 | 6 | 0 | 0 | 6 | 1.00 | 1.00 | 1.00 |
| Task Transfer Function (TTF) | 4 | 0 | 0 | 65 | 0 | 69 | 0.97 | 0.94 | 0.96 |
| Tapered Section (TaS) | 5 | 0 | 0 | 0 | 108 | 113 | 0.98 | 0.96 | 0.97 |
| | AGP | NPS | OoP | TTF | TaS | TOT = 347 | \multicolumn{3}{c}{Accuracy = 0.96} |

**Test Set (Triplicate)**

| | \multicolumn{5}{c}{Confusion Matrix} | N Samples | Precision | Recall | F1 |
|---|---|---|---|---|---|---|---|---|---|
| | \multicolumn{5}{c}{Prediction} | | | | |
| Air Gap or Partial Volume (AGP) | 63 | 4 | 0 | 3 | 3 | 73 | 0.86 | 0.86 | 0.86 |
| Noise Power Spectrum (NPS) | 6 | 78 | 0 | 0 | 0 | 84 | 0.95 | 0.93 | 0.94 |
| Outside Phantom (OoP) | 0 | 0 | 3 | 0 | 0 | 3 | 1.00 | 1.00 | 1.00 |
| Task Transfer Function (TTF) | 2 | 0 | 0 | 66 | 0 | 68 | 0.96 | 0.97 | 0.96 |
| Tapered Section (TaS) | 2 | 0 | 0 | 0 | 113 | 115 | 0.97 | 0.98 | 0.98 |
| | AGP | NPS | OoP | TTF | TaS | TOT = 343 | \multicolumn{3}{c}{Accuracy = 0.94} |

**Extreme Test Set (Triplicate)**

| | \multicolumn{5}{c}{Confusion Matrix} | N Samples | Precision | Recall | F1 |
|---|---|---|---|---|---|---|---|---|---|
| | \multicolumn{5}{c}{Prediction} | | | | |
| Air Gap or Partial Volume (AGP) | 147 | 25 | 0 | 23 | 10 | 205 | 0.92 | 0.72 | 0.81 |
| Noise Power Spectrum (NPS) | 8 | 71 | 0 | 0 | 0 | 79 | 0.66 | 0.90 | 0.76 |
| Outside Phantom (OoP) | 0 | 0 | 4 | 0 | 0 | 4 | 1.00 | 1.00 | 1.00 |
| Task Transfer Function (TTF) | 2 | 0 | 0 | 56 | 0 | 58 | 0.71 | 0.97 | 0.82 |
| Tapered Section (TaS) | 2 | 12 | 0 | 0 | 98 | 112 | 0.91 | 0.88 | 0.89 |
| | AGP | NPS | OoP | TTF | TaS | TOT = 458 | \multicolumn{3}{c}{Accuracy = 0.82} |

Table 3. Confusion matrix and per-class metrics evaluated by classifying the evaluation sets with the triplicate model. Triplicating the input, rather than providing context images, produces a classifier with an identical parameter count but less axial information. Lacking that information slightly increases errors in the properly imaged validation and test sets. There are more errors in classifying images that do or do not have air gaps or partial volumes, but no new type of error. In the extreme test set, we see difficulty in distinguishing tapered sections (TaS) from noise power spectrum patterns (NPS). Axial information is key in detecting this difference, which can be indicated by more subtle associations such as the order of test patterns.

### 3.2.3 Augmentation and the extreme test set

We postulate that little variability arises from the phantoms, which are designed to serve as calibration patterns for CT systems. To verify this claim, in the noAug model augmentation is disabled during training. Plotting the accuracy and loss on the validation set after each epoch (Figure 9), the noAug performance is remarkably smoother as there is fewer variation to learn, and the unaugmented training set is highly similar to the validation set. There appears to be slight overfitting to the unaugmented set, as an accuracy maximum is detectible at approximately 100 epochs, and a subsequent dip of approximately 2% accuracy. Evaluated performance after training is comparable to the control model, with classification accuracy on the validation and test sets at 96% and 97% respectively. The training set is representative of Mercury Phantoms imaged in correct alignment, as are the validation and test sets.

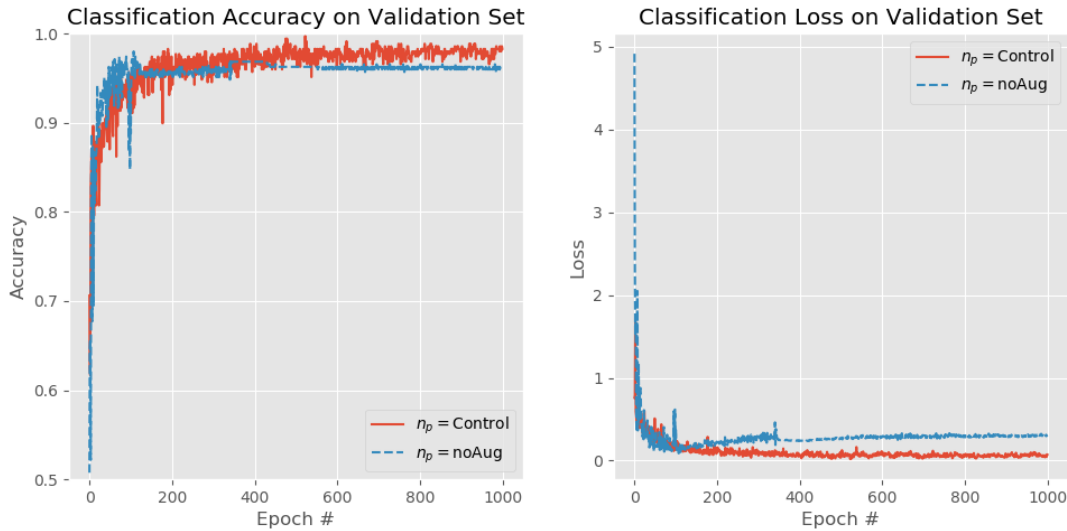

Figure 8. The effect of augmentation on training performance, depicted by plotting model accuracy and loss on validation set evaluated at the end of each training epoch. Accuracy is similar without augmentation, though slightly diminished. There is a distinct smoothness to validation accuracy from one epoch to another, as very little variation remains in the training set without augmentation. The classifier begins to overfit a standard series of mercury phantoms.

When evaluated against misaligned phantom sets, there is a dramatic drop in accuracy. Table 4 shows the previous classification results on the extreme test set with a model trained with augmentation (Aug) as a control, alongside classification results for the non-augmented model. Without augmentation, the model does not recognize patterns in the misaligned phantom, failing to correctly classify any outside phantom (OoP) or tapered section (TaS) images. The out of phantom label is omitted entirely.

TTF patterns are most clearly identified, with the highest overall recall and F1 score. From recall, we see that 91% of TTF patterns are correctly identified. An inability to identify unsuitable qualities in the images causes 20 unsuitable air gap or partial volume (AGP) slices to be classified as TTF patterns, causing a drop in precision for that category. On examination these appear to be unsuitable TTF images, suggesting that the model cannot distinguish TTF suitability when the phantom is misaligned.

The model classifies most images as NPS patterns. Many AGP slices may be unsuitable NPS patterns, however the ratio of AGP vs NGP classifications follows roughly with the prevalence of those patterns, implying no clear distinction. The noAug model does not clearly discriminate between these two categories when the phantom is misaligned. Furthermore, all TaS slices are incorrectly classified as NPS patterns. TaS and NPS segments are composed of the same material, and should be similarly radiopaque when imaged by a CT system. When imaged at an oblique angle, NPS slices may appear slightly oval, as the NPS cylinder is now intersected by the imaging plane at a slanted angle. A slight slope seen at the edge of an NPS pattern may therefore be indistinguishable from the slight slope expected in a tapered section. Without experiencing these artifacts at training time, the model classifies all TaS slices as NPS slices.

Overall the model is unable to classify misaligned phantoms without training on augmented samples. While performance on standard phantom images is comparable to the previous result, a model trained without augmentation is sensitive to imaging conditions and phantom alignment. It does not generalize to the misaligned case.

**Extreme Test Set (Aug)**

| | Confusion Matrix | | | | | N Samples | Precision | Recall | F1 |
|---|---|---|---|---|---|---|---|---|---|
| | | | Prediction | | | | | | |
| Air Gap or Partial Volume (AGP) | 173 | 11 | 0 | 19 | 2 | 205 | 0.84 | 0.84 | 0.84 |
| Noise Power Spectrum (NPS) | 13 | 66 | 0 | 0 | 0 | 79 | 0.86 | 0.84 | 0.85 |
| Outside Phantom (OoP) | 2 | 0 | 2 | 0 | 0 | 4 | 1.00 | 0.50 | 0.67 |
| Task Transfer Function (TTF) | 0 | 0 | 0 | 58 | 0 | 58 | 0.75 | 1.00 | 0.86 |
| Tapered Section (TaS) | 19 | 0 | 0 | 0 | 93 | 112 | 0.98 | 0.83 | 0.90 |
| | AGP | NPS | OoP | TTF | TaS | TOT = 458 | Accuracy = 0.86 | | |

**Extreme Test Set (no Aug)**

| | Confusion Matrix | | | | | N Samples | Precision | Recall | F1 |
|---|---|---|---|---|---|---|---|---|---|
| | | | Prediction | | | | | | |
| Air Gap or Partial Volume (AGP) | 61 | 121 | 0 | 20 | 3 | 205 | 0.78 | 0.30 | 0.43 |
| Noise Power Spectrum (NPS) | 17 | 62 | 0 | 0 | 0 | 79 | 0.20 | 0.78 | 0.32 |
| Outside Phantom (OoP) | 0 | 3 | 0 | 0 | 1 | 4 | 0.00 | 0.00 | 0.00 |
| Task Transfer Function (TTF) | 0 | 5 | 0 | 53 | 0 | 58 | 0.73 | 0.91 | 0.81 |
| Tapered Section (TaS) | 0 | 112 | 0 | 0 | 0 | 112 | 0.00 | 0.00 | 0.00 |
| | AGP | NPS | OoP | TTF | TaS | TOT = 458 | Accuracy = 0.38 | | |

Table 4. Confusion matrices and metrics for the control model, trained with augmentation (Aug), and the experimental variant trained without augmentation (no Aug). There is a pronounced drop in accuracy in classifying this set when no augmentation is used to regularize the phantom classifier. After an equivalent training period, the noAug variant is unable to classify the extreme test set accurately, even underperforming a trivial system that guesses the most frequent category.

## 4 DISCUSSION

The need for manual annotations is one barrier to wider use of quality control instruments such as the Mercury phantom. Automated test pattern annotation greatly reduces the time spent labeling images and standardizes those labels. In our experience this problem has eluded a solution through more traditional image processing methods due largely to the subtle and difficult-to-articulate features that make an image unsuitable for measurement. An operator requires some expertise to consistently select the ideal images for estimating CT system imaging characteristics. The deep learning approach presented here agrees with expert annotation in a majority of cases and is robust to improper positioning, even tolerating the exaggerated misplacement in the extreme test set.

A limitation of this study was the time required for manual annotations. Large libraries of images on these phantoms are available at medical centers that regularly conduct these quality assurance measurements. The 3,557 images in total used in this study are a relatively small data set compared to standards such as ImageNet but require a significant reader effort to label. Multiple readers would facilitate the use of a larger data set while allowing some evaluation of inter- and intra-reader variability. Given the small margin of error achieved by this model on standard data sets, it is worth investigating whether experts agree as frequently as 98%, or if the model emulates the original reader too closely. If the readings are repeatable, the model parameterized here may be effective for automatically labeling larger data sets for further model training and evaluation via semi-supervised learning.

Although we were able to train several computationally intensive models, we still imposed some limitations on the variety of parameter modifications and metrics reported. Wherever possible, training parameters were consistent, and metrics were simplified to enable direct comparisons as often as possible. We acknowledge that in some situations a model variant may have been further optimized. For instance, the variation without augmentation may have benefitted from early stopping or some other form of reduced training to avoid overfitting[13]. That particular case would complicate comparisons by swapping one form of regularization for another. Similarly, the accuracy deficit in the random initialization model's evaluation of the extreme test set may have been overcome with additional training. We also do not adjust other training hyperparameters, such as the choice of optimizer, associated parameters, loss function, or batch size[19,20]. No study can truly encompass all the available training variations, rather we choose a subset that may give insight into what information the model is incorporating.

The resulting model does accurately classify phantom images in ideal settings and retains most of that accuracy in more difficult conditions. It is fascinating that although as much as 88% of the total model parameters are in the dense classification layers in the $n_p = 4096$ case, these classification layers do not provide much of the accuracy. A much smaller model with $n_p = 256$ achieves similar accuracy in all evaluation sets at approximately 1/6 the model size. This finding is consistent with initial tests of VGG classifiers, which attributed much of the architecture's reported accuracy to the convolutional layers[12].

Another interesting finding is that initial layers trained for interpreting color channels require little to no modification to interpret additional image information. Our classifier performed comparably well when forced to retain ImageNet weights at early layers. There were definite uses of the axial information for a subset of classifications. This is evidenced by how the accuracy suffers and new modes of failure arise in the triplicate model, which lacks axial information. Given that freezing the lower layers does not notably affect accuracy, and likewise no accuracy losses are observed with substantially lower parameter counts in the top classification layers, we conclude that axial associations are made in the mid-to-upper convolutional layers. Research into neural network activation and interpretability suggests similar findings [7,21].

We proposed that misalignment and other imaging parameters introduce variability which can be emulated through extensive augmentation of well-aligned sets, rather than collecting samples with misalignment. Regularization by image augmentation was critical in training this model to evaluate misaligned phantoms. While the augmentations were imperfect emulations of real changes to the images, they imparted crucial properties like scale invariance to the classifier. Removing augmentation had the most pronounced effect on classification accuracy of any of the ablation studies, including triplicating input. In future studies it may be worth reiterating on augmentations to more closely match artifacts native to CT systems or phantom misuse.

Ultimately deploying this model in more conventional clinical hardware may benefit from smaller architectures with suitable accuracy but lower computational requirements. VGG19 was useful in this study for its straightforward architecture and ease of training. Various alternative classification architectures exist which may improve performance or reduce resources requirements. For example, architectures such as ResNet[6], Inception[3,5,16], or Inception-Resnet[5] provide demonstrable performance gains over VGG classifiers. MobileNet[22] and its variations reduce the total computational requirement, at times with negligible performance penalties. The ablation experiments performed here show ImageNet weights do have demonstrable benefits in training, but other random initialization schemes may allow for training novel architectures for this problem with no performance penalty. Testing all these tradeoffs when applied to CT phantom images are the next logical steps in single-image classification approach.

## 5 CONCLUSIONS

These results indicate an effective approach to automating this facet of image quality assurance. We achieve excellent performance in independent data sets from typical imaging conditions for phantoms, matching expert annotations in as much as 98% of cases. The classifier is insensitive to even gross faults in positioning the phantom, achieving 86% accuracy in the extreme test set. Image augmentation, and the reuse of input channels trained for color images, were effective in adapting the VGG19 architecture to this domain.

# APPENDIX

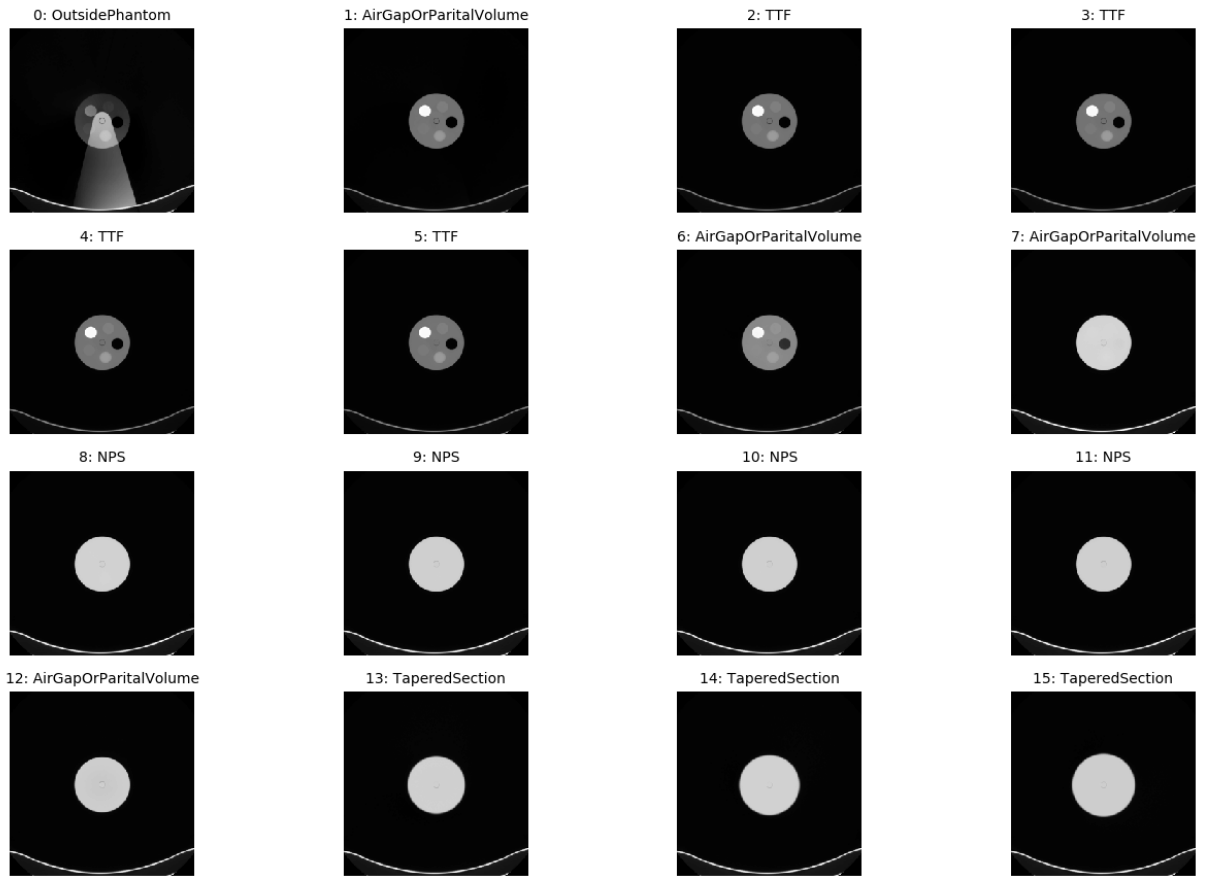

Figure A1. A series of Mercury Phantom images from the training set. These images are in order of acquisition, without augmentation or shuffling. A clinical imaging physicist acquired each phantom image series and labeled each image with the entire series for reference. Note the position relative to the CT bed at the bottom of the frame. The first image shows part of the supporting structure shown in Figure 1. These are the smallest diameter images, coming from the beginning of the phantom.

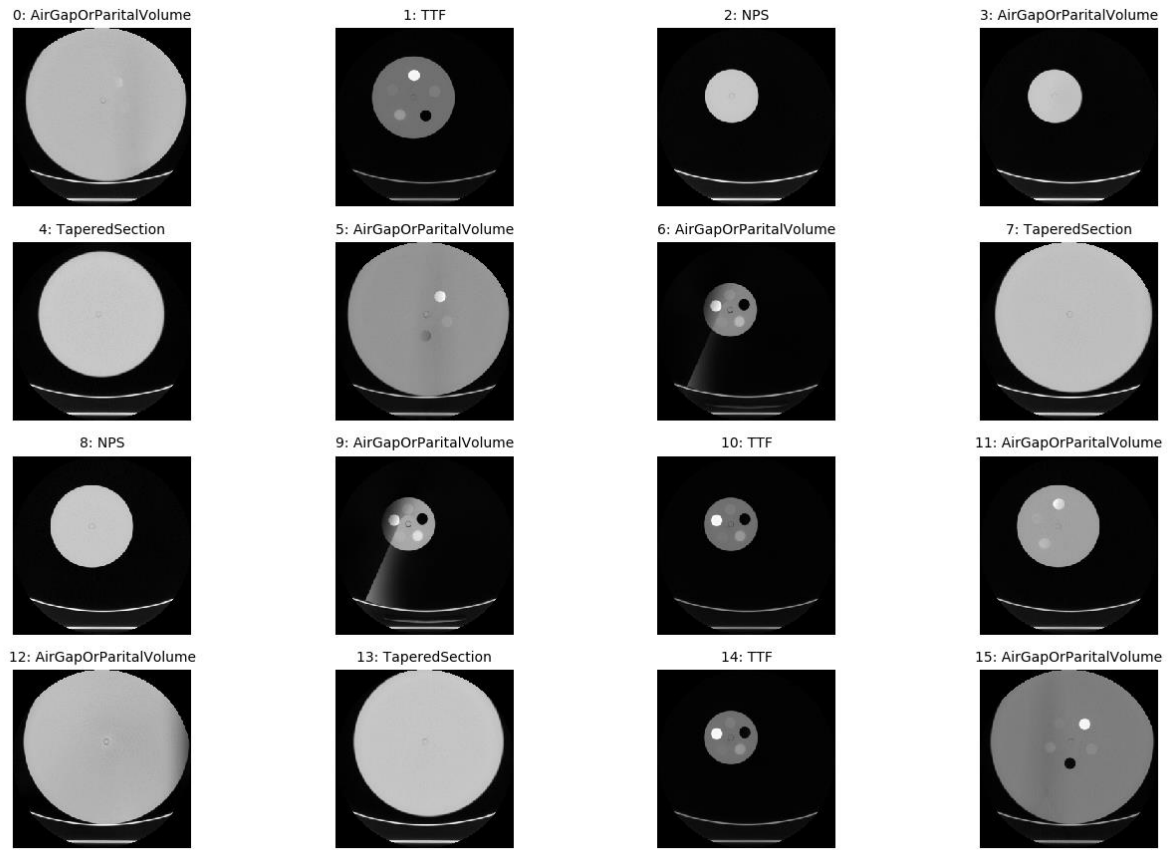

Figure A2. A series of Mercury Phantom images from the extreme test set. These images are in shuffled order, with no augmentation. Compared to the training set, the field of view does not always fully encompass the phantom, causing cutoff artifacts as seen in images 0, 5, 7, 12, 13, and 15. The support structure is partly visible in images 6 and 9, and is placed off center with respect to the CT bed. All of these intentional misalignments also notably shift the smaller diameter images off-center, such as in image 2. It is difficult to misalign the phantom dramatically and still acquire an image, as the physical device must pass through the CT bore. Several images in this series are graded as unsuitable (Air gap or partial volume) by the clinical imaging physicist labeling the phantom images.

**Validation Set (nP = 256)**

Confusion Matrix

| Truth \ Prediction | AGP | NPS | OoP | TTF | TaS | N Samples | Precision | Recall | F1 |
|---|---|---|---|---|---|---|---|---|---|
| Air Gap or Partial Volume (AGP) | 71 | 0 | 0 | 2 | 0 | 73 | 0.95 | 0.97 | 0.96 |
| Noise Power Spectrum (NPS) | 2 | 84 | 0 | 0 | 0 | 86 | 1.00 | 0.98 | 0.99 |
| Outside Phantom (OoP) | 0 | 0 | 6 | 0 | 0 | 6 | 1.00 | 1.00 | 1.00 |
| Task Transfer Function (TTF) | 2 | 0 | 0 | 67 | 0 | 69 | 0.97 | 0.97 | 0.97 |
| Tapered Section (TaS) | 0 | 0 | 0 | 0 | 113 | 113 | 1.00 | 1.00 | 1.00 |
| | AGP | NPS | OoP | TTF | TaS | TOT = 347 | Accuracy = 0.98 | | |

**Validation Set (nP = 512)**

Confusion Matrix

| Truth \ Prediction | AGP | NPS | OoP | TTF | TaS | N Samples | Precision | Recall | F1 |
|---|---|---|---|---|---|---|---|---|---|
| Air Gap or Partial Volume (AGP) | 73 | 0 | 0 | 0 | 0 | 73 | 0.92 | 1.00 | 0.96 |
| Noise Power Spectrum (NPS) | 3 | 83 | 0 | 0 | 0 | 86 | 1.00 | 0.97 | 0.98 |
| Outside Phantom (OoP) | 0 | 0 | 6 | 0 | 0 | 6 | 1.00 | 1.00 | 1.00 |
| Task Transfer Function (TTF) | 2 | 0 | 0 | 67 | 0 | 69 | 1.00 | 0.97 | 0.99 |
| Tapered Section (TaS) | 1 | 0 | 0 | 0 | 112 | 113 | 1.00 | 0.99 | 1.00 |
| | AGP | NPS | OoP | TTF | TaS | TOT = 347 | Accuracy = 0.98 | | |

**Validation Set (nP = 1024)**

Confusion Matrix

| Truth \ Prediction | AGP | NPS | OoP | TTF | TaS | N Samples | Precision | Recall | F1 |
|---|---|---|---|---|---|---|---|---|---|
| Air Gap or Partial Volume (AGP) | 73 | 0 | 0 | 0 | 0 | 73 | 0.91 | 1.00 | 0.95 |
| Noise Power Spectrum (NPS) | 4 | 82 | 0 | 0 | 0 | 86 | 1.00 | 0.95 | 0.98 |
| Outside Phantom (OoP) | 0 | 0 | 6 | 0 | 0 | 6 | 1.00 | 1.00 | 1.00 |
| Task Transfer Function (TTF) | 0 | 0 | 0 | 69 | 0 | 69 | 1.00 | 1.00 | 1.00 |
| Tapered Section (TaS) | 3 | 0 | 0 | 0 | 110 | 113 | 1.00 | 0.97 | 0.99 |
| | AGP | NPS | OoP | TTF | TaS | TOT = 347 | Accuracy = 0.98 | | |

**Validation Set (nP = 2048)**

Confusion Matrix

| Truth \ Prediction | AGP | NPS | OoP | TTF | TaS | N Samples | Precision | Recall | F1 |
|---|---|---|---|---|---|---|---|---|---|
| Air Gap or Partial Volume (AGP) | 73 | 0 | 0 | 0 | 0 | 73 | 0.91 | 1.00 | 0.95 |
| Noise Power Spectrum (NPS) | 2 | 84 | 0 | 0 | 0 | 86 | 1.00 | 0.98 | 0.99 |
| Outside Phantom (OoP) | 0 | 0 | 6 | 0 | 0 | 6 | 1.00 | 1.00 | 1.00 |
| Task Transfer Function (TTF) | 4 | 0 | 0 | 65 | 0 | 69 | 1.00 | 0.94 | 0.97 |
| Tapered Section (TaS) | 1 | 0 | 0 | 0 | 112 | 113 | 1.00 | 0.99 | 1.00 |
| | AGP | NPS | OoP | TTF | TaS | TOT = 347 | Accuracy = 0.98 | | |

**Validation Set (nP = 4096)**

Confusion Matrix

| Truth \ Prediction | AGP | NPS | OoP | TTF | TaS | N Samples | Precision | Recall | F1 |
|---|---|---|---|---|---|---|---|---|---|
| Air Gap or Partial Volume (AGP) | 72 | 0 | 0 | 1 | 0 | 73 | 0.90 | 0.99 | 0.94 |
| Noise Power Spectrum (NPS) | 7 | 79 | 0 | 0 | 0 | 86 | 0.99 | 0.92 | 0.95 |
| Outside Phantom (OoP) | 0 | 0 | 6 | 0 | 0 | 6 | 1.00 | 1.00 | 1.00 |
| Task Transfer Function (TTF) | 0 | 0 | 0 | 69 | 0 | 69 | 0.99 | 1.00 | 0.99 |
| Tapered Section (TaS) | 1 | 1 | 0 | 0 | 111 | 113 | 1.00 | 0.98 | 0.99 |
| | AGP | NPS | OoP | TTF | TaS | TOT = 347 | Accuracy = 0.97 | | |

Table A1. Confusion matrices and per-category metrics for the model in Figure 3 evaluated on the validation set after training. Results are reported for each variant as the number of channels in the dense layer ($n_p$/nP) is increased by factors of 2. All model variants classify the validation set with high accuracy. With the exception of one tapered section misclassification by the 4096 variant, these models only mistake the presence or absence of air gaps or partial volumes that make images unsuitable, in a minority of cases. The models are overall indistinguishable by validation set accuracy.

**Test Set (nP = 256)**

| Truth | Confusion Matrix | | | | | N Samples | Precision | Recall | F1 |
|---|---|---|---|---|---|---|---|---|---|
| | | | Prediction | | | | | | |
| Air Gap or Partial Volume (AGP) | 71 | 0 | 0 | 2 | 0 | 73 | 0.96 | 0.97 | 0.97 |
| Noise Power Spectrum (NPS) | 1 | 83 | 0 | 0 | 0 | 84 | 1.00 | 0.99 | 0.99 |
| Outside Phantom (OoP) | 0 | 0 | 3 | 0 | 0 | 3 | 1.00 | 1.00 | 1.00 |
| Task Transfer Function (TTF) | 1 | 0 | 0 | 67 | 0 | 68 | 0.97 | 0.99 | 0.98 |
| Tapered Section (TaS) | 1 | 0 | 0 | 0 | 114 | 115 | 1.00 | 0.99 | 1.00 |
| | AGP | NPS | OoP | TTF | TaS | TOT = 343 | Accuracy = 0.99 | | |

**Test Set (nP = 512)**

| Truth | Confusion Matrix | | | | | N Samples | Precision | Recall | F1 |
|---|---|---|---|---|---|---|---|---|---|
| | | | Prediction | | | | | | |
| Air Gap or Partial Volume (AGP) | 71 | 0 | 0 | 2 | 0 | 73 | 0.90 | 0.97 | 0.93 |
| Noise Power Spectrum (NPS) | 3 | 81 | 0 | 0 | 0 | 84 | 1.00 | 0.96 | 0.98 |
| Outside Phantom (OoP) | 0 | 0 | 3 | 0 | 0 | 3 | 0.75 | 1.00 | 0.86 |
| Task Transfer Function (TTF) | 3 | 0 | 0 | 65 | 0 | 68 | 0.97 | 0.96 | 0.96 |
| Tapered Section (TaS) | 2 | 0 | 1 | 0 | 112 | 115 | 1.00 | 0.97 | 0.99 |
| | AGP | NPS | OoP | TTF | TaS | TOT = 343 | Accuracy = 0.97 | | |

**Test Set (nP = 1024)**

| Truth | Confusion Matrix | | | | | N Samples | Precision | Recall | F1 |
|---|---|---|---|---|---|---|---|---|---|
| | | | Prediction | | | | | | |
| Air Gap or Partial Volume (AGP) | 70 | 0 | 0 | 2 | 1 | 73 | 0.91 | 0.96 | 0.93 |
| Noise Power Spectrum (NPS) | 4 | 80 | 0 | 0 | 0 | 84 | 1.00 | 0.95 | 0.98 |
| Outside Phantom (OoP) | 0 | 0 | 3 | 0 | 0 | 3 | 1.00 | 1.00 | 1.00 |
| Task Transfer Function (TTF) | 2 | 0 | 0 | 66 | 0 | 68 | 0.97 | 0.97 | 0.97 |
| Tapered Section (TaS) | 1 | 0 | 0 | 0 | 114 | 115 | 0.99 | 0.99 | 0.99 |
| | AGP | NPS | OoP | TTF | TaS | TOT = 343 | Accuracy = 0.97 | | |

**Test Set (nP = 2048)**

| Truth | Confusion Matrix | | | | | N Samples | Precision | Recall | F1 |
|---|---|---|---|---|---|---|---|---|---|
| | | | Prediction | | | | | | |
| Air Gap or Partial Volume (AGP) | 72 | 0 | 0 | 1 | 0 | 73 | 0.87 | 0.99 | 0.92 |
| Noise Power Spectrum (NPS) | 5 | 79 | 0 | 0 | 0 | 84 | 1.00 | 0.94 | 0.97 |
| Outside Phantom (OoP) | 0 | 0 | 3 | 0 | 0 | 3 | 1.00 | 1.00 | 1.00 |
| Task Transfer Function (TTF) | 4 | 0 | 0 | 64 | 0 | 68 | 0.98 | 0.94 | 0.96 |
| Tapered Section (TaS) | 2 | 0 | 0 | 0 | 113 | 115 | 1.00 | 0.98 | 0.99 |
| | AGP | NPS | OoP | TTF | TaS | TOT = 343 | Accuracy = 0.97 | | |

**Test Set (nP = 4096)**

| Truth | Confusion Matrix | | | | | N Samples | Precision | Recall | F1 |
|---|---|---|---|---|---|---|---|---|---|
| | | | Prediction | | | | | | |
| Air Gap or Partial Volume (AGP) | 71 | 0 | 0 | 2 | 0 | 73 | 0.88 | 0.97 | 0.92 |
| Noise Power Spectrum (NPS) | 7 | 77 | 0 | 0 | 0 | 84 | 1.00 | 0.92 | 0.96 |
| Outside Phantom (OoP) | 0 | 0 | 3 | 0 | 0 | 3 | 1.00 | 1.00 | 1.00 |
| Task Transfer Function (TTF) | 1 | 0 | 0 | 67 | 0 | 68 | 0.97 | 0.99 | 0.98 |
| Tapered Section (TaS) | 2 | 0 | 0 | 0 | 113 | 115 | 1.00 | 0.98 | 0.99 |
| | AGP | NPS | OoP | TTF | TaS | TOT = 343 | Accuracy = 0.97 | | |

Table A2. Confusion matrices and per-category metrics for the model in Figure 3 evaluated on the test set after training. Results are reported for each variant as the number of channels in the dense layer ($n_p$/nP) is increased by factors of 2. Similar to validation set results, all model variants classify the test set with high accuracy. One tapered section image is misclassified as an Out of Phantom image, but otherwise errors are in determining suitable images. The models are overall indistinguishable by test set accuracy, which are classified correctly in most cases.

**Extreme Test Set (nP = 256)**

Confusion Matrix

| Truth \ Prediction | AGP | NPS | OoP | TTF | TaS | N Samples | Precision | Recall | F1 |
|---|---|---|---|---|---|---|---|---|---|
| Air Gap or Partial Volume (AGP) | 156 | 10 | 5 | 30 | 4 | 205 | 0.89 | 0.76 | 0.82 |
| Noise Power Spectrum (NPS) | 7 | 72 | 0 | 0 | 0 | 79 | 0.87 | 0.91 | 0.89 |
| Outside Phantom (OoP) | 4 | 0 | 0 | 0 | 0 | 4 | 0.00 | 0.00 | 0.00 |
| Task Transfer Function (TTF) | 0 | 0 | 0 | 58 | 0 | 58 | 0.66 | 1.00 | 0.79 |
| Tapered Section (TaS) | 8 | 1 | 0 | 0 | 103 | 112 | 0.96 | 0.92 | 0.94 |
| | AGP | NPS | OoP | TTF | TaS | TOT = 458 | Accuracy = 0.85 | | |

**Extreme Test Set (nP = 512)**

| Truth \ Prediction | AGP | NPS | OoP | TTF | TaS | N Samples | Precision | Recall | F1 |
|---|---|---|---|---|---|---|---|---|---|
| Air Gap or Partial Volume (AGP) | 173 | 11 | 0 | 19 | 2 | 205 | 0.84 | 0.84 | 0.84 |
| Noise Power Spectrum (NPS) | 13 | 66 | 0 | 0 | 0 | 79 | 0.86 | 0.84 | 0.85 |
| Outside Phantom (OoP) | 2 | 0 | 2 | 0 | 0 | 4 | 1.00 | 0.50 | 0.67 |
| Task Transfer Function (TTF) | 0 | 0 | 0 | 58 | 0 | 58 | 0.75 | 1.00 | 0.86 |
| Tapered Section (TaS) | 19 | 0 | 0 | 0 | 93 | 112 | 0.98 | 0.83 | 0.90 |
| | AGP | NPS | OoP | TTF | TaS | TOT = 458 | Accuracy = 0.86 | | |

**Extreme Test Set (nP = 1024)**

| Truth \ Prediction | AGP | NPS | OoP | TTF | TaS | N Samples | Precision | Recall | F1 |
|---|---|---|---|---|---|---|---|---|---|
| Air Gap or Partial Volume (AGP) | 153 | 18 | 2 | 22 | 10 | 205 | 0.78 | 0.75 | 0.76 |
| Noise Power Spectrum (NPS) | 8 | 71 | 0 | 0 | 0 | 79 | 0.79 | 0.90 | 0.84 |
| Outside Phantom (OoP) | 4 | 0 | 0 | 0 | 0 | 4 | 0.00 | 0.00 | 0.00 |
| Task Transfer Function (TTF) | 0 | 0 | 0 | 58 | 0 | 58 | 0.72 | 1.00 | 0.84 |
| Tapered Section (TaS) | 31 | 1 | 0 | 0 | 80 | 112 | 0.89 | 0.71 | 0.79 |
| | AGP | NPS | OoP | TTF | TaS | TOT = 458 | Accuracy = 0.79 | | |

**Extreme Test Set (nP = 2048)**

| Truth \ Prediction | AGP | NPS | OoP | TTF | TaS | N Samples | Precision | Recall | F1 |
|---|---|---|---|---|---|---|---|---|---|
| Air Gap or Partial Volume (AGP) | 155 | 24 | 0 | 21 | 5 | 205 | 0.85 | 0.76 | 0.80 |
| Noise Power Spectrum (NPS) | 9 | 70 | 0 | 0 | 0 | 79 | 0.62 | 0.89 | 0.73 |
| Outside Phantom (OoP) | 4 | 0 | 0 | 0 | 0 | 4 | 0.00 | 0.00 | 0.00 |
| Task Transfer Function (TTF) | 2 | 0 | 0 | 56 | 0 | 58 | 0.73 | 0.97 | 0.83 |
| Tapered Section (TaS) | 12 | 18 | 0 | 0 | 82 | 112 | 0.94 | 0.73 | 0.82 |
| | AGP | NPS | OoP | TTF | TaS | TOT = 458 | Accuracy = 0.79 | | |

**Extreme Test Set (nP = 4096)**

| Truth \ Prediction | AGP | NPS | OoP | TTF | TaS | N Samples | Precision | Recall | F1 |
|---|---|---|---|---|---|---|---|---|---|
| Air Gap or Partial Volume (AGP) | 181 | 2 | 1 | 13 | 8 | 205 | 0.82 | 0.88 | 0.85 |
| Noise Power Spectrum (NPS) | 24 | 55 | 0 | 0 | 0 | 79 | 0.96 | 0.70 | 0.81 |
| Outside Phantom (OoP) | 4 | 0 | 0 | 0 | 0 | 4 | 0.00 | 0.00 | 0.00 |
| Task Transfer Function (TTF) | 3 | 0 | 0 | 55 | 0 | 58 | 0.81 | 0.95 | 0.87 |
| Tapered Section (TaS) | 8 | 0 | 0 | 0 | 104 | 112 | 0.93 | 0.93 | 0.93 |
| | AGP | NPS | OoP | TTF | TaS | TOT = 458 | Accuracy = 0.86 | | |

Table A3. Confusion matrices and per-category metrics for the model in Figure 3 evaluated on the extreme test set after training. Results are reported for each variant as the number of channels in the dense layer ($n_p$/nP) is increased by factors of 2. The more difficult cases present in the extreme test set help differentiate the model variants. The 1024 and 2048 variants underperform in this evaluation. While the 256 variant dramatically reduces the size of the classifier at similar overall accuracy, the types of errors it produces are undesirable in practice. By contrast, the 512 and 4096 variants are conservative, and more likely to classify an image as unsuitable rather than classify an unsuitable AGP image as a valid test pattern. We select the 512 model as it minimizes the overall parameter count while achieving good accuracy with more favorable types of error in the extreme test evaluation.